\pdfoutput=1

\documentclass[11pt]{article}

\usepackage[]{EMNLP2024}

\usepackage{times}
\usepackage{latexsym}

\usepackage[T1]{fontenc}

\usepackage[utf8]{inputenc}

\usepackage{microtype}

\usepackage[scaled]{beramono}

\usepackage{xcolor}
\usepackage[T1]{fontenc}
\usepackage{color-edits}
\usepackage{textcomp}
\usepackage{listings}
\usepackage{lipsum}

\addauthor{MM}{purple} 

\definecolor{codegreen}{rgb}{0,0.6,0}
\definecolor{codegray}{rgb}{0.5,0.5,0.5}
\definecolor{codepurple}{rgb}{0.58,0,0.82}
\definecolor{backcolour}{rgb}{0.95,0.95,0.92}

\lstdefinestyle{code_style}{
    backgroundcolor=\color{backcolour},   
    commentstyle=\color{codegreen},
    keywordstyle=\color{magenta},
    numberstyle=\tiny\color{codegray},
    numbers=none,
    upquote=true,
    stringstyle=\color{codepurple},
    basicstyle=\ttfamily~\footnotesize,
    breakatwhitespace=false,         
    breaklines=false,                 
    captionpos=b,                    
    keepspaces=true,                 
    numbers=left,                    
    numbersep=5pt,                  
    showspaces=false,                
    showstringspaces=false,
    showtabs=false,                  
    tabsize=2,
}

\usepackage{graphicx,scalerel,xparse}

\NewDocumentCommand\hfemoji{}{
    \scalerel*{
        \includegraphics{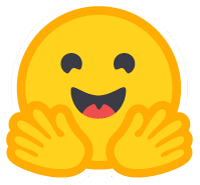}
    }{X}
}
\title{\hfemoji AutoTrain:\\No-code training for state-of-the-art models}

\author{Abhishek Thakur \\
   Hugging Face, Inc. \\
  \texttt{abhishek@huggingface.co}
}

\begin{document}
\maketitle
\begin{abstract}
With the advancements in open-source models, training(or finetuning) models on custom datasets has become a crucial part of developing solutions which are tailored to specific industrial or open-source applications. Yet, there is no single tool which simplifies the process of training across different types of modalities or tasks.
We introduce~\hfemoji \emph{AutoTrain} (aka AutoTrain Advanced)---an open-source, no code tool/library which can be used to train (or finetune) models for different kinds of tasks such as: large language model (LLM) finetuning, text classification/regression, token classification, sequence-to-sequence task, finetuning of sentence transformers, visual language model (VLM) finetuning, image classification/regression and even classification and regression tasks on tabular data.~\hfemoji  \emph{AutoTrain Advanced} is an open-source library providing best practices for training models on custom datasets. The library is available at \href{https://github.com/huggingface/autotrain-advanced}{https://github.com/huggingface/autotrain-advanced}. AutoTrain can be used in fully local mode or on cloud machines and works with tens of thousands of models shared on Hugging Face Hub and their variations.

\textbf{Demo screencast:} \href{https://youtu.be/2O0jHC99S0k}{YouTube}

\end{abstract}


\section{Introduction}
\label{sec:introduction}

With recent advancements in open-source and open-access state-of-the-art models, the need for standardized yet customizable training of models on downstream tasks has become crucial. However, a universal open-source solution for a diverse range of tasks is still lacking. To address this challenge, we introduce \hfemoji \emph{AutoTrain} (also known as AutoTrain Advanced).

AutoTrain is an open-source solution which offers model training for different kinds of tasks such as: large language model (LLM) finetuning, text classification/scoring, token classification, training custom embedding models using sentence transformers \cite{reimers-2019-sentence-bert}, finetuning for visual language models (VLMs), computer vision tasks such as image classification/scoring, object detection and even tabular regression and classification tasks. At the time of writing this paper, a total of 22 tasks: 16 text-based, 4 image-based and 2 tabular based have been implemented.

\begin{figure*}[t]
    \centering
    \includegraphics[width=1.0\textwidth]{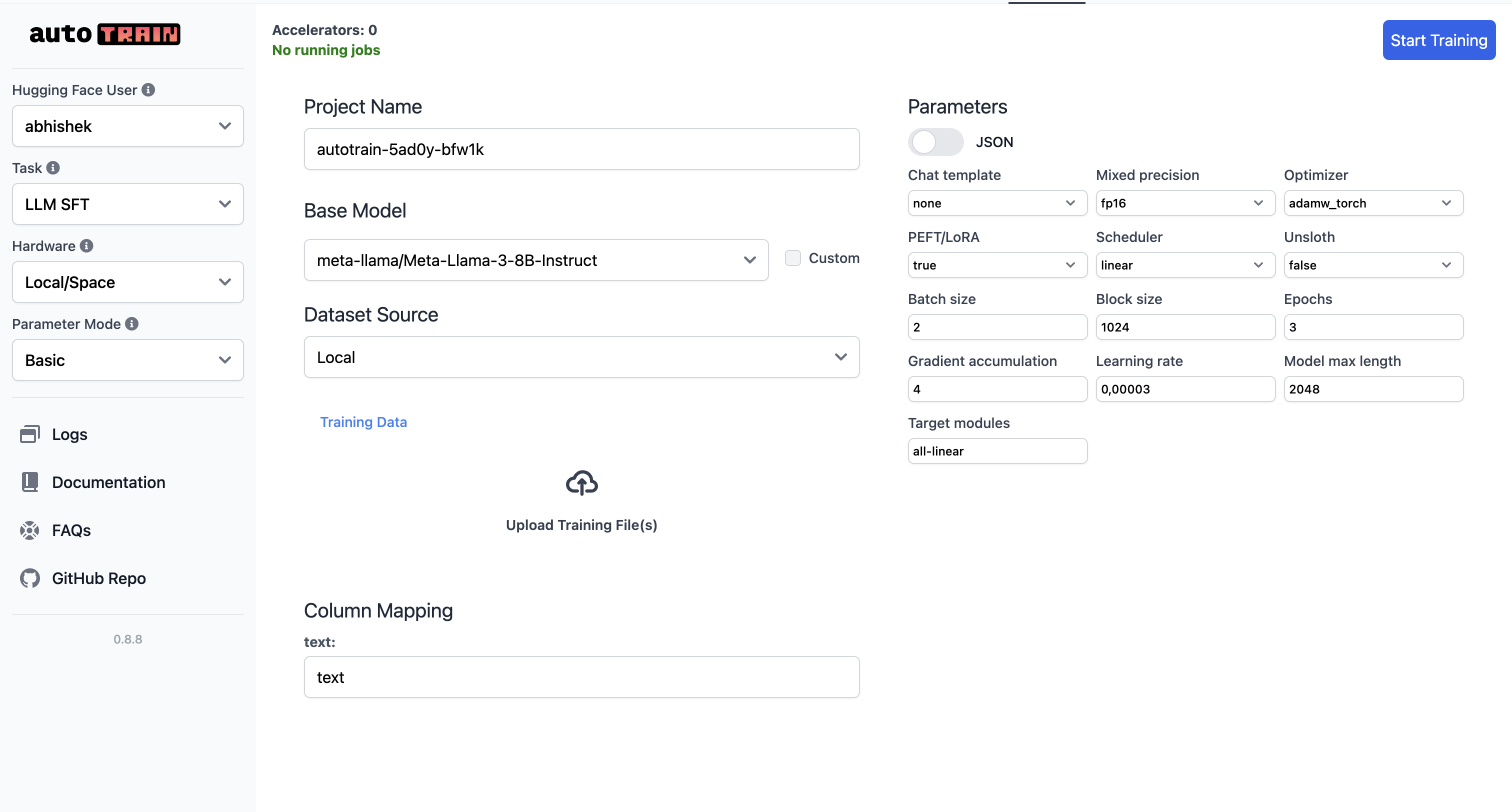}
    \caption{A screenshot of the AutoTrain User Interface (UI)}
    \label{fig:autotrain_ui}
\end{figure*}

The idea behind creating AutoTrain is to allow a simple interface for training models on custom datasets without requiring extensive knowledge of coding. AutoTrain is intended for not just no-coders but also for experienced data scientists and machine learning practioners. Instead of writing complex scripts, one can focus on gathering and preparing your data and let AutoTrain handle the training part. AutoTrain UI is shown in Figure~\ref{fig:autotrain_ui}.

When talking about model training, there are several problems which arise:

\textit{Complexity of hyperparameter tuning}: Finding the right parameters for tuning models can only be done by significant experimentations and expertise. Improperly tuning the hyperparameters can result in overfitting or underfitting.

\textit{Model validation}: A good way to make sure the trained models generalize well, is to have a proper valiation set and a proper way to evaluate with appropriate metrics. Overfitting to training data can cause the models to fail in real-world scenarios.

\textit{Distributed training}: Training models on larger datasets with multi-gpu support can be cumbersome and requires significant changes to codebase. Distributed training requires additional complexity when it comes to synchronization and data handling.

\textit{Monitoring}: While training a model, its crucial to monitor losses, metrics and artifacts to make sure there is nothing fishy going on.

\textit{Maintenance}: With ever-changing data, it may be necessary to retrain or fine-tune the model on new data while keeping the training settings consistent.

We introduce the open source \emph{AutoTrain Advanced} library to address many of these problems. 

\section{Related work}

In recent years, many AutoML solutions have been developed to automate the training process of machine learning models. Some notable solutions include:

\paragraph{AutoML Solutions} 

AutoSklearn \cite{feurer-neurips15a}, which is an open-source AutoML toolkit built on top of the popular scikit-learn library. AutoSklearn uses Bayesian optimization to automate the process of model selection and hyperparameter tuning. 

AutoCompete \cite{thakur2015autocompeteframeworkmachinelearning}, which won the Codalab AutoML GPU challenges, builds a framework for tacking machine learning competitions. The code is, however, not open source.

Axolotl \cite{axolotl} is a CLI tool for finetuning LLMs.

AutoKeras \cite{JMLR:v24:20-1355}, developed on top of Keras offers functionalities for various tasks such as image classification, text classification, and regression

Many other closed-source solutions have also been developed by Google, Microsoft, and Amazon. However, all these solutions have some limitations. They are either not open-source, or if they are, they can only handle a limited number of tasks. Many of these solutions are also not no-code, making them inaccessible to non-coders.

With \textit{AutoTrain}, we provide a single interface to deal with many different data format, task, and model combinations, which depending on user's choices is also closely connnected to Hugging Face Hub which enables download, inference and sharing of models with the entire world. Moreover, AutoTrain supports almost all kinds models which are compatible with Hugging Face Transformers \cite{wolf2019huggingface} library, making it a unique solution to support hundreds of thousands of models for finetuning, including the models which require custom code.

\section{Library: \textit{AutoTrain Advanced}}

The \emph{AutoTrain Advanced} python library provides a command line interface (CLI), a graphical user interface (GUI/UI) and python SDK to enable training on custom datasets. The datasets can be uploaded/used in different formats such as zip files, CSVs or JSONLs. We provide documentation and walkthroughs on training models for different task and dataset combinations with example hyperparameters, evaluation results and usage of the trained models. The library is licensed as Apache 2.0 license and is available on Github,~\footnote{\href{github.com/huggingface/autotrain-advanced}{https://github.com/huggingface/autotrain-advanced}} making it easy for anyone to adopt and contribute.

The design of the library has been made keeping in mind both professionals and amateurs who would like to finetune model but don't know where to start and don't want to invest time setting up a separate environment for each of their finetuning tasks. The library lies on the shoulders of giants such as Transformers \cite{wolf2019huggingface}, Hugging Face Datasets \cite{lhoest2021datasets}, Accelerate \cite{accelerate}, Diffusers\cite{von-platen-etal-2022-diffusers}, PEFT \cite{peft}, TRL \cite{vonwerra2022trl} and other libraries created by Hugging Face.

AutoTrain uses \cite{paszke2019pytorch} as the main backend for training the models. For tabular datasets, models from \cite{van2014scikit} and \cite{Chen_2016} are used as preferred models.

\subsection{Component of the AutoTrain Advanced library}
There are 3 main components in the \emph{AutoTrain Advanced} library:

\paragraph{Project Configuration:} manages the configuration of the project and allows users to set up and manage their training projects. Here, one can specify various settings such as the type of task (e.g., llm finetuning, text classification, image classification), dataset, the model to use, and other training parameters. This step ensures that all necessary configurations are in place before starting the training process.

\paragraph{Dataset Processor:} handles the preparation and preprocessing of datasets. It ensures that data is in the right format for training. This component can handle different types of data, including text, images, and tabular data. Dataset processor does cleaning and transformation of dataset, saves time and reduces the potential for errors. A dataset once processed can also be used for multiple projects without requiring to be processed again.

\paragraph{Trainer:} is responsible for the actual training process. It manages the training loop, handles the computation of loss and metrics, and optimizes the model. The Trainer also supports distributed training, allowing you to train models on multiple GPUs seamlessly. Additionally, it includes tools for monitoring the training progress, ensuring that everything is running smoothly and efficiently.

\subsection{Installation \& Usage}

Using \emph{AutoTrain} is as easy as pie. In this section we focus briefly on installation and LLM finetuning task. However, the same can be applied to other tasks keeping in mind the dataset format which is provided

\paragraph{Installation} AutoTrain Advanced can be easily installed using pip. 

\begin{lstlisting}[language=bash, numbers=none]
$ pip install autotrain-advanced
\end{lstlisting}

It has to be noted that the the pip installation doesnt install pytorch and users must install it on their own. However, a complete package with all the requirements is also available as a docker image.

\begin{lstlisting}[language=bash, numbers=none]
$ docker pull
   huggingface/autotrain-advanced:latest
\end{lstlisting}

\begin{figure*}[t]
    \centering
    \includegraphics[width=1.0\textwidth]{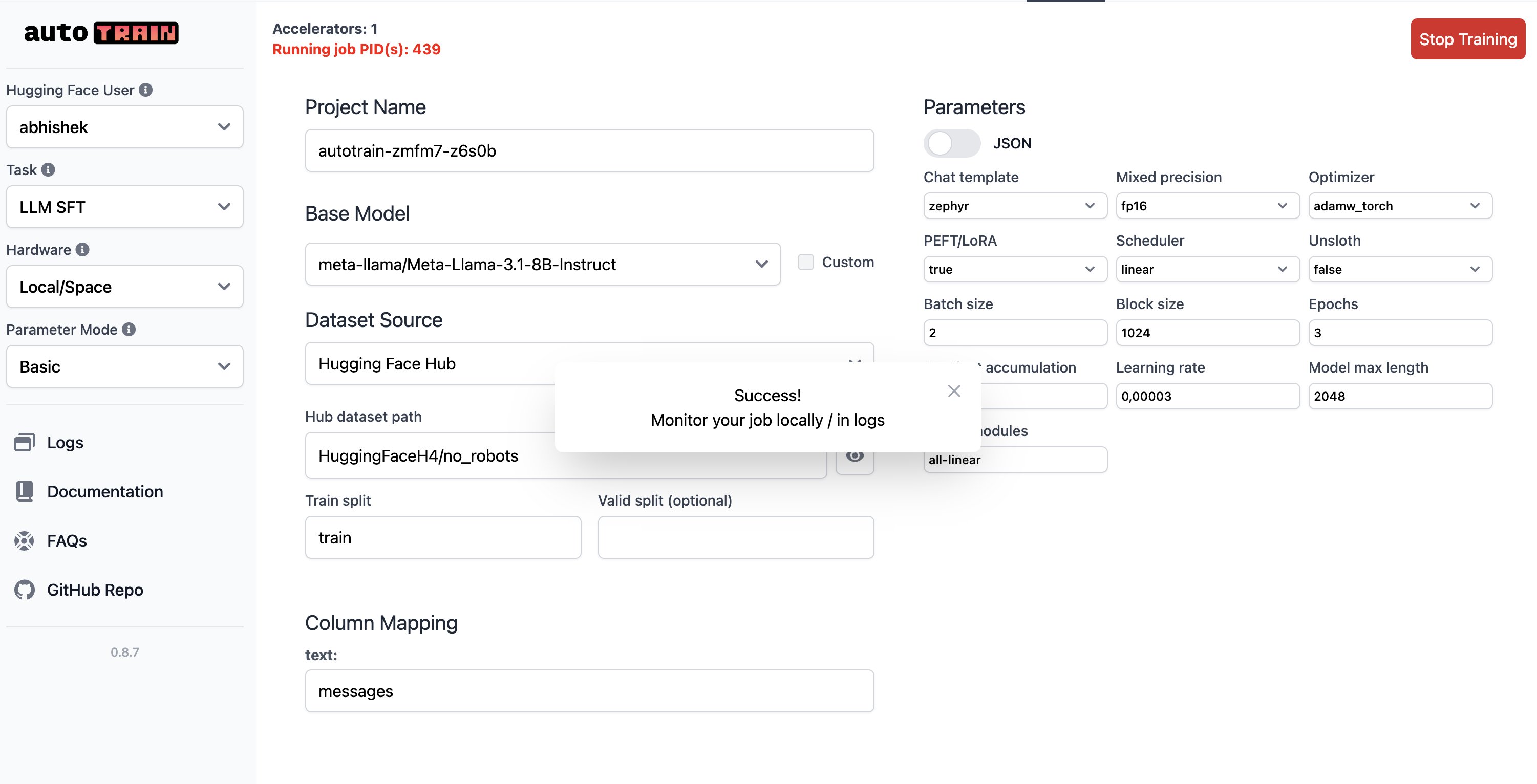}
    \caption{Finetuing an LLM in AutoTrain UI}
    \label{fig:autotrain_ui_training}
\end{figure*}

\paragraph{Usage} AutoTrain Advanced offers CLI and UI. CLI is based on a AutoTrain Advanced python library. So, users familiar with python can also use the python sdk. To start the UI as shown in Figure~\ref{fig:autotrain_ui}, one can run the \texttt{autotrain app} command:

\begin{lstlisting}[language=bash, numbers=none]
$ autotrain app
\end{lstlisting}

An example of running training in UI is shown in Figure~\ref{fig:autotrain_ui_training}.

Training can also be started using a config file which is in yaml format and the autotrain cli. An example config to finetune llama 3.1 is shown below:

\begin{lstlisting}
task: llm:orpo
base_model: meta-llama/Meta-Llama-3.1-8B
project_name: autotrain-llama
log: tensorboard
backend: local

data:
  path: HuggingFaceH4/no_robots
  train_split: train
  valid_split: null
  chat_template: zephyr
  column_mapping:
    text_column: chosen
    rejected_text_column: rejected
    prompt_text_column: prompt

params:
  block_size: 1024
  model_max_length: 8192
  max_prompt_length: 512
  epochs: 3
  batch_size: 2
  lr: 3e-5
  peft: true
  quantization: int4
  target_modules: all-linear
  padding: right
  optimizer: adamw_torch
  scheduler: linear
  gradient_accumulation: 4
  mixed_precision: fp16

hub:
  username: ${HF_USERNAME}
  token: ${HF_TOKEN}
  push_to_hub: true
\end{lstlisting}

The above config file shows how a \texttt{llama-3.1-8B} model from the Hugging Face hub can be finetuned on \texttt{HuggingFaceH4/no\_robots} dataset which is also available on Hugging Face Hub. If the user wants to use a local dataset and model, they can do that too by following the documentation. In this specific case, a local dataset can be provided as a JSONL file. To start the training, \texttt{autotrain --config} command is used:

\begin{lstlisting}[language=bash, numbers=none]
$ autotrain --config config.yml
\end{lstlisting}

The training process starts tensorboard \cite{tensorflow2015-whitepaper} which can be used to monitor the training and metrics and generated during the training process. The users can also monitor the training logs in terminal if they started the training using the CLI or in the UI logs section. 

The trained model, depending on user's choice, can also be pushed to Hugging Face Hub, thus making it accessible to hundreds of thousands of users across the world. The trained models are also compatible with major inference providers (huggingface, aws, google cloud, etc.) which makes deployment and consumption easy for both coders and non-coders.

\section{Conclusion}

In this paper, we introduce \emph{AutoTrain} (aka AutoTrain Advanced), which is an open source, no-code solution for training (or finetuning) machine learning models on a variety of tasks. AutoTrain addresses common challenges in the model training process, such as dataset processing, hyperparameter tuning, model validation, distributed training, monitoring, and maintenance. By automating these tasks, AutoTrain ensures that users can efficiently build high-performing models without needing extensive coding knowledge or experience. Additionally, AutoTrain supports a diverse range of tasks, including llm finetuning, text classification, image classification, and regression, and even tabular data classification/regression, thus, making it a versatile tool for various applications.

\section*{Limitations}
AutoTrain tries to generalize the training process for a given model - dataset combination as much as possible, however, there might be situations in which custom changes might be required. For example, AutoTrain doesnt provide support for sample weights, model merging, or ensembling yet. We are gathering issues faced by users and implementing them to address these limitations.



\section*{Acknowledgements}

We thank the many contributors to the Hugging Face open source ecosystem. We also thank the different teams at Hugging Face: the open-source team, the infrastructure team, the hub team, frontend and backend teams and others.

\bibliography{custom}

\begin{thebibliography}{15}
\expandafter\ifx\csname natexlab\endcsname\relax\def\natexlab#1{#1}\fi

\bibitem[{Abadi et~al.(2015)Abadi, Agarwal, Barham, Brevdo, Chen, Citro, Corrado, Davis, Dean, Devin, Ghemawat, Goodfellow, Harp, Irving, Isard, Jia, Jozefowicz, Kaiser, Kudlur, Levenberg, Man\'{e}, Monga, Moore, Murray, Olah, Schuster, Shlens, Steiner, Sutskever, Talwar, Tucker, Vanhoucke, Vasudevan, Vi\'{e}gas, Vinyals, Warden, Wattenberg, Wicke, Yu, and Zheng}]{tensorflow2015-whitepaper}
Mart\'{i}n Abadi, Ashish Agarwal, Paul Barham, Eugene Brevdo, Zhifeng Chen, Craig Citro, Greg~S. Corrado, Andy Davis, Jeffrey Dean, Matthieu Devin, Sanjay Ghemawat, Ian Goodfellow, Andrew Harp, Geoffrey Irving, Michael Isard, Yangqing Jia, Rafal Jozefowicz, Lukasz Kaiser, Manjunath Kudlur, Josh Levenberg, Dandelion Man\'{e}, Rajat Monga, Sherry Moore, Derek Murray, Chris Olah, Mike Schuster, Jonathon Shlens, Benoit Steiner, Ilya Sutskever, Kunal Talwar, Paul Tucker, Vincent Vanhoucke, Vijay Vasudevan, Fernanda Vi\'{e}gas, Oriol Vinyals, Pete Warden, Martin Wattenberg, Martin Wicke, Yuan Yu, and Xiaoqiang Zheng. 2015.
\newblock \href {https://www.tensorflow.org/} {{TensorFlow}: Large-scale machine learning on heterogeneous systems}.
\newblock Software available from tensorflow.org.

\bibitem[{Chen and Guestrin(2016)}]{Chen_2016}
Tianqi Chen and Carlos Guestrin. 2016.
\newblock \href {https://doi.org/10.1145/2939672.2939785} {Xgboost: A scalable tree boosting system}.
\newblock In \emph{Proceedings of the 22nd ACM SIGKDD International Conference on Knowledge Discovery and Data Mining}, KDD ’16. ACM.

\bibitem[{Cloud(2024)}]{axolotl}
Axolotl~AI Cloud. 2024.
\newblock Axolotl: A tool for streamlining fine-tuning of ai models.
\newblock \url{https://github.com/axolotl-ai-cloud/axolotl}.
\newblock Accessed: 2024-08-06.

\bibitem[{Feurer et~al.(2015)Feurer, Klein, Eggensperger, Springenberg, Blum, and Hutter}]{feurer-neurips15a}
Matthias Feurer, Aaron Klein, Katharina Eggensperger, Jost Springenberg, Manuel Blum, and Frank Hutter. 2015.
\newblock Efficient and robust automated machine learning.
\newblock In \emph{Advances in Neural Information Processing Systems 28 (2015)}, pages 2962--2970.

\bibitem[{Gugger et~al.(2022)Gugger, Debut, Wolf, Schmid, Mueller, Mangrulkar, Sun, and Bossan}]{accelerate}
Sylvain Gugger, Lysandre Debut, Thomas Wolf, Philipp Schmid, Zachary Mueller, Sourab Mangrulkar, Marc Sun, and Benjamin Bossan. 2022.
\newblock Accelerate: Training and inference at scale made simple, efficient and adaptable.
\newblock \url{https://github.com/huggingface/accelerate}.

\bibitem[{Jin et~al.(2023)Jin, Chollet, Song, and Hu}]{JMLR:v24:20-1355}
Haifeng Jin, François Chollet, Qingquan Song, and Xia Hu. 2023.
\newblock \href {http://jmlr.org/papers/v24/20-1355.html} {Autokeras: An automl library for deep learning}.
\newblock \emph{Journal of Machine Learning Research}, 24(6):1--6.

\bibitem[{Lhoest et~al.(2021)Lhoest, del Moral, Jernite, Thakur, von Platen, Patil, Chaumond, Drame, Plu, Tunstall et~al.}]{lhoest2021datasets}
Quentin Lhoest, Albert~Villanova del Moral, Yacine Jernite, Abhishek Thakur, Patrick von Platen, Suraj Patil, Julien Chaumond, Mariama Drame, Julien Plu, Lewis Tunstall, et~al. 2021.
\newblock \href {https://aclanthology.org/2021.emnlp-demo.21/} {Datasets: A community library for natural language processing}.
\newblock \emph{arXiv preprint arXiv:2109.02846}.

\bibitem[{Mangrulkar et~al.(2022)Mangrulkar, Gugger, Debut, Belkada, Paul, and Bossan}]{peft}
Sourab Mangrulkar, Sylvain Gugger, Lysandre Debut, Younes Belkada, Sayak Paul, and Benjamin Bossan. 2022.
\newblock Peft: State-of-the-art parameter-efficient fine-tuning methods.
\newblock \url{https://github.com/huggingface/peft}.

\bibitem[{Paszke et~al.(2019)Paszke, Gross, Massa, Lerer, Bradbury, Chanan, Killeen, Lin, Gimelshein, Antiga et~al.}]{paszke2019pytorch}
Adam Paszke, Sam Gross, Francisco Massa, Adam Lerer, James Bradbury, Gregory Chanan, Trevor Killeen, Zeming Lin, Natalia Gimelshein, Luca Antiga, et~al. 2019.
\newblock \href {https://proceedings.neurips.cc/paper/2019/file/bdbca288fee7f92f2bfa9f7012727740-Paper.pdf} {Pytorch: An imperative style, high-performance deep learning library}.
\newblock \emph{Advances in neural information processing systems}, 32.

\bibitem[{Reimers and Gurevych(2019)}]{reimers-2019-sentence-bert}
Nils Reimers and Iryna Gurevych. 2019.
\newblock \href {https://arxiv.org/abs/1908.10084} {Sentence-bert: Sentence embeddings using siamese bert-networks}.
\newblock In \emph{Proceedings of the 2019 Conference on Empirical Methods in Natural Language Processing}. Association for Computational Linguistics.

\bibitem[{Thakur and Krohn-Grimberghe(2015)}]{thakur2015autocompeteframeworkmachinelearning}
Abhishek Thakur and Artus Krohn-Grimberghe. 2015.
\newblock \href {http://arxiv.org/abs/1507.02188} {Autocompete: A framework for machine learning competition}.

\bibitem[{Van~der Walt et~al.(2014)Van~der Walt, Sch{\"o}nberger, Nunez-Iglesias, Boulogne, Warner, Yager, Gouillart, and Yu}]{van2014scikit}
Stefan Van~der Walt, Johannes~L Sch{\"o}nberger, Juan Nunez-Iglesias, Fran{\c{c}}ois Boulogne, Joshua~D Warner, Neil Yager, Emmanuelle Gouillart, and Tony Yu. 2014.
\newblock \href {https://peerj.com/articles/453/?report=reader&utm_source=TrendMD&utm_campaign=PeerJ_TrendMD_1&utm_medium=TrendMD} {scikit-image: image processing in python}.
\newblock \emph{PeerJ}, 2:e453.

\bibitem[{von Platen et~al.(2022)von Platen, Patil, Lozhkov, Cuenca, Lambert, Rasul, Davaadorj, Nair, Paul, Berman, Xu, Liu, and Wolf}]{von-platen-etal-2022-diffusers}
Patrick von Platen, Suraj Patil, Anton Lozhkov, Pedro Cuenca, Nathan Lambert, Kashif Rasul, Mishig Davaadorj, Dhruv Nair, Sayak Paul, William Berman, Yiyi Xu, Steven Liu, and Thomas Wolf. 2022.
\newblock Diffusers: State-of-the-art diffusion models.
\newblock \url{https://github.com/huggingface/diffusers}.

\bibitem[{von Werra et~al.(2020)von Werra, Belkada, Tunstall, Beeching, Thrush, Lambert, and Huang}]{vonwerra2022trl}
Leandro von Werra, Younes Belkada, Lewis Tunstall, Edward Beeching, Tristan Thrush, Nathan Lambert, and Shengyi Huang. 2020.
\newblock Trl: Transformer reinforcement learning.
\newblock \url{https://github.com/huggingface/trl}.

\bibitem[{Wolf et~al.(2019)Wolf, Debut, Sanh, Chaumond, Delangue, Moi, Cistac, Rault, Louf, Funtowicz et~al.}]{wolf2019huggingface}
Thomas Wolf, Lysandre Debut, Victor Sanh, Julien Chaumond, Clement Delangue, Anthony Moi, Pierric Cistac, Tim Rault, R{\'e}mi Louf, Morgan Funtowicz, et~al. 2019.
\newblock Huggingface's transformers: State-of-the-art natural language processing.
\newblock \emph{arXiv preprint arXiv:1910.03771}.

\end{thebibliography}
\bibliographystyle{acl_natbib}

\end{document}